\documentclass[runningheads]{llncs}
\usepackage[T1]{fontenc}
\usepackage{booktabs}

\usepackage{graphicx}
\usepackage{hyperref}

\usepackage{color}

\usepackage{graphicx,verbatim}
\usepackage{booktabs}
\usepackage{pifont}
\usepackage{multirow}
\usepackage{hyperref}
\newcommand{\cmark}{\ding{51}}
\newcommand{\xmark}{\ding{55}}

\usepackage{microtype}
\begin{document}
\title{MIS-VQA: An Annotation-Driven Surgical Vision–Language Dataset for Complex Surgical Understanding}
%
\titlerunning{Annotation-Driven Surgical Vision--Language Modeling}
%
%
\author{Mohamed Hamdy\inst{1} \and
Iheb Zouari\inst{1} \and
Fatmaelzahraa Ahmed\inst{2} \and
Muraam Abdel-Ghani\inst{2} \and
Muhammed Arsalan\inst{1} \and
Ponnuthurai Suganthan\inst{1} \and
Khalid Al-Jalham\inst{2} \and
Abdulaziz Al-Ali\inst{1} \and
Shidin Balakrishnan\thanks{Corresponding author: SBalakrishnan1@hamad.qa} \inst{2} 
}
\authorrunning{M. Hamdy et al .}
%
\institute{Department of Computer Science and Engineering, Qatar University, Doha, Qatar
\and
Hamad Medical Corporation, Doha, Qatar\\
}
\maketitle              
\begin{abstract}
Modern surgical AI increasingly requires multimodal understanding rather than isolated perception, yet existing datasets remain largely limited to single-task supervision such as recognition, segmentation, or templated question answering. Current medical VQA resources often rely on synthetic templates, lack natural descriptive supervision, or provide sparse annotations that hinder reasoning about spatial relations and procedural context. We introduce MIS-VQA, a surgical vision–language dataset generated through an annotation-driven pipeline that converts heterogeneous surgical annotations into structured textual scene representations. Using EndoVis18 and CholecT50, we transform phase, instrument, interaction, triplet, and segmentation labels into unified contexts that guide large language models to generate clinically coherent question–answer pairs. MIS-VQA supports four response types—concise and detailed descriptions, perception-level grounding, and multi-step reasoning—enabling hierarchical evaluation from atomic visual grounding to compositional reasoning. To address annotation sparsity, we incorporate confidence-filtered segmentation predictions and cross-annotation consistency checks, enabling spatially grounded supervision at scale. MIS-VQA establishes a unified benchmark for surgical vision–language understanding with spatial supervision and multi-tier reasoning evaluation. To demonstrate its utility, we introduce LLaVA-SurgXpert, a family of surgical vision–language models with interchangeable visual encoders to analyze how visual representations influence multimodal reasoning. Experiments show consistent improvements over existing surgical VQA baselines while revealing capability gaps between perceptual grounding and higher-level reasoning hidden in prior benchmarks. Together, MIS-VQA and LLaVA-SurgXpert establish annotation-driven data generation and structured evaluation as scalable foundations for multimodal surgical AI. 
The dataset and implementation are made available on GitHub\footnote{\url{https://github.com/M-Hamdy-M/MIS-VQA}}.



\end{abstract}

\begin{figure}
\includegraphics[width=\textwidth]{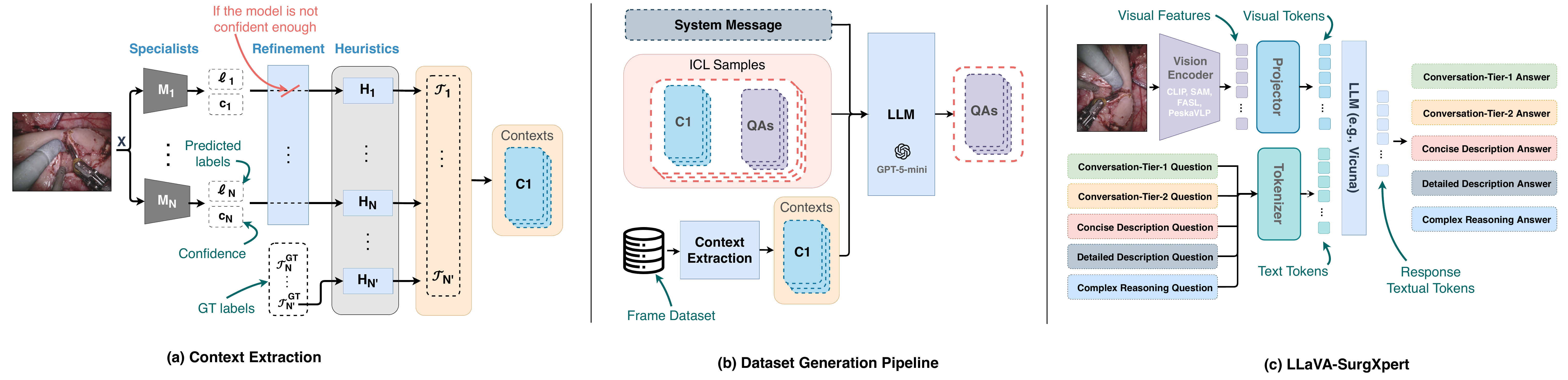}
\caption{Overview of the proposed framework. (a) Context extraction from multi-source annotations using refined specialist model predictions and heuristics for textual abstraction. (b) Structured dataset generation via system messages and expert-curated ICL exemplars. (c) LLaVA-SurgXpert architecture for unified captioning, grounding, and multimodal reasoning in MIS.} \label{fig:pipeline}
\end{figure}

\section{Introduction}
Minimally Invasive Surgery (MIS) has transformed modern surgical practice by reducing trauma, shortening recovery time, and improving procedural precision  \cite{lancet}. The widespread adoption of laparoscopic and robotic platforms has simultaneously led to the large-scale availability of surgical video data, catalyzing rapid progress in surgical data science \cite{laparoscopic_review,cv_surgery}. Recent advances in surgical data science have enabled substantial progress in task-specific computer vision problems such as phase recognition \cite{endonet}, tool-tissue interaction analysis \cite{inst_tissue1,inst_tissue2}, segmentation \cite{Deep_Learning_Sys,endovis18,cholecseg8k}, and action triplet recognition \cite{rendezvous_triplet}. While these systems have achieved strong performance within their respective domains, they operate in isolation and offer limited support for flexible, language-based interaction. In contrast, recent progress in multimodal large language models (MLLMs) and visual instruction tuning~\cite{llava} has demonstrated the potential to unify perception and reasoning within a single framework, enabling models to interpret visual scenes and respond to diverse natural language queries. Early attempts at surgical VQA and multimodal dialogue have shown initial feasibility; however, they are typically limited by small-scale datasets, rigid template-driven question design, incomplete exploitation of spatial annotations, and a lack of structured separation between perceptual grounding and higher-level reasoning--gaps that hinder comprehensive evaluation and scalable development of clinically reliable multimodal systems.

A key barrier to surgical vision-language modeling is the limited depth and structure of existing supervision. Most surgical VQA datasets focus on short factual queries or object presence classification~\cite{endovis_vqa}, providing little support for compositional reasoning involving spatial relations, instrument interactions, and procedural context. Spatially grounded questions are often constrained by the availability of dense annotations, restricting broader scene-level evaluation \cite{endochat}. Moreover, several datasets rely on template-based generation~\cite{ssg_vqa}, producing rigid linguistic patterns and reducing natural question diversity. Captioning supervision--particularly both concise and detailed scene descriptions--remains inconsistent across datasets \cite{endochat}, preventing comprehensive evaluation of captioning, grounding, and reasoning within a unified benchmark.

To address these limitations, we introduce \textbf{MIS-VQA}, a large-scale surgical vision-language dataset constructed through a structured annotation-to-language pipeline. Given the difficulty of directly extracting fine-grained surgical semantics from raw images, we convert multi-source annotations--including segmentation masks, phase labels, instrument identities, and interaction triplets--into standardized textual scene representations that serve as structured context for question-answer generation. Expert-designed in-context learning (ICL) exemplars, curated by surgeons, ensure clinically coherent language, consistent terminology, and realistic question formulation. MIS-VQA defines four response types spanning concise and detailed captioning, complex reasoning, and conversational evaluation, the latter decomposed into atomic visual grounding at the entity level and compositional reasoning across instruments, anatomies, and spatial relations. This hierarchical design enables unified benchmarking of captioning, grounding, and higher-level multimodal reasoning in minimally invasive surgery.

To demonstrate its utility, we further introduce \textbf{LLaVA-SurgXpert}, a family of surgical vision-language models built upon a unified alignment framework with interchangeable visual encoders. By comparing general-purpose, task-oriented, and domain-specific pretrained representations, we analyze how visual priors influence spatial grounding and reasoning performance in operative environments. Together, MIS-VQA and LLaVA-SurgXpert establish a scalable framework for annotation-driven surgical vision-language learning. Our main contributions are: 1) a structured annotation-to-language framework for generating clinically grounded surgical VQA data; 2) MIS-VQA, a hierarchical benchmark unifying captioning, grounding, and reasoning evaluation; and 3) LLaVA-SurgXpert, a controlled experimental framework for analyzing the impact of visual encoder pretraining on multimodal reasoning in minimally invasive surgery.

\section{Methodology}
\subsection{MIS-VQA: Dataset Construction}
\begin{figure}[b]
\includegraphics[width=\textwidth]{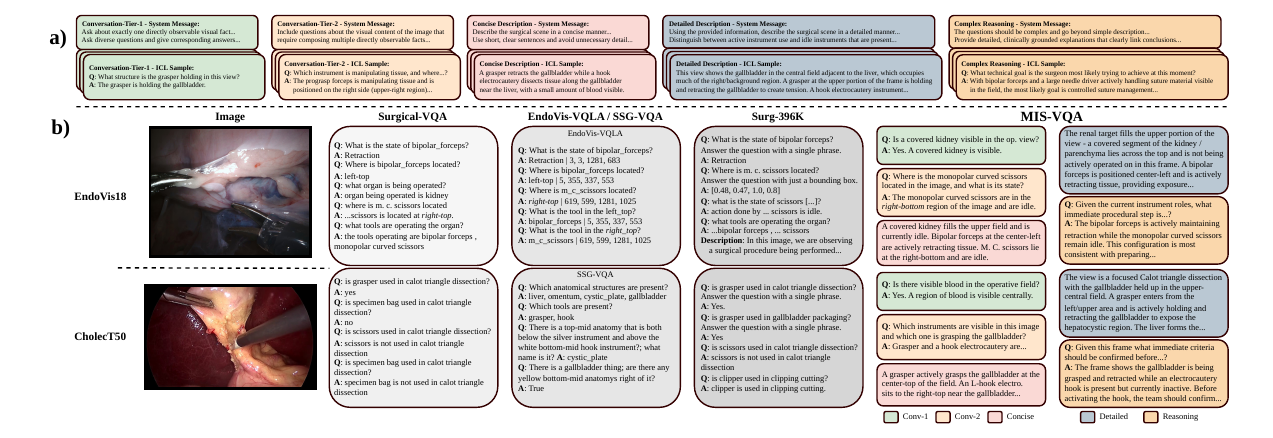}
\caption{Overview of the generated data. (a) Excerpt of the system messages and representative ICL samples used for MIS-VQA data generation. (b) Samples from the generated MIS-VQA data in comparison with existing datasets.} \label{fig:icl}
\end{figure}

MIS-VQA is constructed through a structured annotation-to-language pipeline inspired by visual instruction tuning~\cite{llava}, but adapted to the surgical domain where dense natural captions are unavailable. Unlike conventional visual instruction tuning pipelines that rely on captions and bounding boxes as symbolic surrogates for images, we leverage clinically grounded multi-source surgical annotations converted into standardized textual scene representations as the primary source of semantic context. For each image, available annotations are transformed into a unified textual description describing visible entities, spatial relations, and procedural state. This context is provided to an LLM, which is instructed to generate question–answer pairs as if directly observing the image, as illustrated in Fig.~\ref{fig:pipeline}a,b. 

To ensure clinical realism and linguistic consistency, we design dataset-specific system prompts and curate expert-written in-context learning (ICL) exemplars (illustrative examples shown in Fig.~\ref{fig:icl}). These ICL samples, created by surgeons, consist of extracted annotation contexts paired with manually written questions, answers, and captions. MIS-VQA defines four response types: (1) concise description for clinically efficient summarization, (2) detailed description for comprehensive scene understanding, (3) complex reasoning requiring multi-entity compositional interpretation, and (4) conversational evaluation, which is hierarchically decomposed into Tier-1 atomic visual grounding and Tier-2 compositional scene reasoning. This hierarchical design enables targeted evaluation of perceptual grounding versus structured reasoning capabilities.

For EndoVis18, two complementary annotation sources are available for all frames: semantic segmentation masks \cite{endovis18} and instrument–tissue interaction labels \cite{inst_tissue1}. Since ground-truth annotations are complete, no auxiliary specialist models are required. Segmentation masks provide spatial grounding of instruments and anatomical structures, while interaction labels encode fine-grained action semantics. These annotations are directly converted into structured textual context following the heuristics introduced in \cite{hamdy2026decoupling}, ensuring reliable spatial and interaction-aware supervision across all generated response types.

In contrast, CholecT50 contains heterogeneous and partially overlapping annotations. Phase and action triplet labels are obtained from CholecT50~\cite{rendezvous_triplet}, instrument presence from the Cholec80 (Cholec45 intersection)~\cite{endonet}, and segmentation masks from the limited CholecSeg8k subset~\cite{cholecseg8k}. Because segmentation masks are sparsely available, we train the FASL surgical segmentation model~\cite{fasl} on the full CholecSeg8k dataset (train+test) to predict masks for the remaining frames. Predicted masks are incorporated only when confidence exceeds 0.9, otherwise spatial context is omitted. The system prompt explicitly informs the LLM whether spatial information originates from ground truth or model predictions, encouraging consistency with more reliable annotations. We observe that this mechanism preserves spatial reasoning while reducing errors caused by noisy segmentation predictions.

To improve dataset quality and diversity, we introduce adaptive frame sampling and annotation refinement. Frames are sampled when at least one of phase, instrument, or action-triplet annotations changes relative to the previous frame. To avoid large temporal gaps, we enforce a maximum skip interval of five seconds (FPS=1) and insert midpoint frames when longer static segments occur. We additionally filter visually invalid frames and remove inconsistent triplets that reference instruments absent from the corresponding instrument annotations. This cross-validation step reduces annotation noise and improves the semantic consistency of generated VQAs. The resulting dataset contains over 800K vision-language pairs with spatially grounded and hierarchically structured supervision.


\subsection{LLaVA-SurgXpert: Surgical Vision-Language Modeling}
Our model follows a standard vision–language alignment framework composed of a visual encoder, a projection layer, and a Vicuna LLM (Fig.~\ref{fig:pipeline}c). The LLM and projector architectures are the same across all variants, while we explore five different visual encoders to study the impact of domain specialization and task priors.

\textbf{Visual encoders.} We categorize encoders into three groups: 
(1) \emph{general-purpose}: CLIP; 
(2) \emph{task-oriented}: SAM; 
(3) \emph{domain-specific surgical models}: FASL and PeskaVLP. 
CLIP provides strong general visual representations, while SAM offers segmentation-aware spatial priors. FASL is optimized for surgical instrument segmentation, and 
PeskaVLP~\cite{peskavlp} is a CLIP-style surgical video-language model with a ResNet-based visual encoder.

\textbf{Token alignment.} Since encoder outputs differ in spatial resolution and channel dimensionality, we standardize them to 1024 visual tokens before projection. 
FASL features are reshaped via pixel unshuffle: $128\!\times\!128\!\times\!64 \rightarrow 32\!\times\!32\!\times\!1024$ (1024 tokens with 1024 channels). 
CLIP follows the standard LLaVA tokenization. 
SAM features are processed by its neck $64\!\times\!64\!\times\!1024 \rightarrow 64\!\times\!64\!\times\!256$, then pixel-unshuffled to $32\!\times\!32\!\times\!1024$. 
For PeskaVLP, the feature map $7\!\times\!7\!\times\!2048$ is rearranged via space-to-depth to $28\!\times\!28\!\times\!128$ and bilinearly interpolated to $32\!\times\!32\!\times\!128$, producing 1024 tokens; channel alignment is handled by the projector.

\textbf{Training strategy.} Training follows a two-stage scheme. 
During \emph{pretraining}, only the projection layer is optimized using concise and detailed description data to align visual and language embeddings. 
During \emph{instruction fine-tuning}, the projector and LoRA adapters of the LLM are trained jointly on all response types, enabling structured reasoning while preserving the base language model.

\section{Experiments and Results}

\subsection{Experimental Setup}

\textbf{LLM Selection for Dataset Generation.}
Given their demonstrated ability to generate clinically coherent and structured surgical descriptions~\cite{hamdy2026decoupling,midlshort}, we evaluate three OpenAI models (GPT-5, GPT-5-mini, and GPT-5-nano) on a representative subset during dataset construction. GPT-5 produces highly detailed but overly verbose responses, while GPT-5-nano tends to generate shorter yet occasionally generic or less reliable outputs. GPT-5-mini achieves the best balance between clinical fidelity, structured reasoning, and conciseness, and is therefore used for full-scale MIS-VQA generation.

\textbf{Datasets and Splits.}
For EndoVis18, we maintain the same training and testing protocol as EndoVis-VQA, using videos \{1, 5, 16\} for evaluation. For CholecT50, since our dataset composition differs from Cholec80-VQA, we preserve the original Cholec80-VQA test intersection, evaluating on videos \{5, 12, 26, 27, 31\}, while all remaining videos are used for training. Dataset statistics, including number of images and samples per response type, are summarized in Table~\ref{tab:dataset_comp_and_stats}.

\textbf{Implementation Details, Baselines, and Evaluation.}
Pretraining is conducted for two epochs with a learning rate of $1\mathrm{e}{-3}$ (cosine schedule), followed by one epoch of instruction fine-tuning at $2\mathrm{e}{-5}$. For mask generation, FASL is trained on the full CholecSeg8k dataset (train+test). When used as a visual encoder in LLaVA-SurgXpert, it is trained exclusively on non-overlapping training videos to prevent data leakage. EndoVis experiments are performed on a single RTX 4090 (batch size 1), while Cholec experiments use 2×H100 GPUs (per-device batch size 16). We train VisualBert \cite{visualbert}, VisualBert-RM \cite{endovis_vqa}, and EndoChat \cite{endochat} using their official training protocols and compare against them. Performance is evaluated using BLEU, CIDEr, METEOR, cosine similarity (text-embedding-3-large), and a normalized GPT-Score. For GPT-Score, GPT-5-nano is conditioned on the annotation-derived context and evaluates predicted answers along visual grounding, clinical correctness, logical consistency, completeness, and hallucination penalty. Ground-truth answers are scored identically for normalization.

\begin{table*}
\centering
\footnotesize

\caption{Comparison of surgical VQA datasets (left) and MIS-VQA statistics (right). 
TQ/TA denote template-based questions/answers; Cap. and Reas. indicates the availability of captioning and complex reasoning questions.}
\label{tab:dataset_comp_and_stats}
\setlength{\tabcolsep}{3pt}
\resizebox{\textwidth}{!}{%
\begin{tabular}{l|ccccccccc||l|cc|cc}
\hline
\multicolumn{10}{c||}{\textbf{Dataset Comparison}} & 
\multicolumn{5}{c}{\textbf{MIS-VQA Statistics}} \\
 &  &  &  &  &  &  &  &  &  
&  & \multicolumn{2}{c}{\# Imgs} & \multicolumn{2}{c}{\# Samples} \\
Dataset & Public$^{*}$ & \#Procedures & \#Imgs & \#Samples & Spatial & TQ & TA & Cap. & Reas.
& Resp. Type & Train & Test & Train & Test \\
\hline
EndoVis-18-VQA \cite{endovis_vqa} & \cmark & 1 & 2K & 11.8K & Endo & \cmark & \cmark & \xmark & \xmark
& Concise & 29K & 3.7K & 29K & 3.7K \\

Cholec-80-VQA \cite{endovis_vqa} & \cmark & 1 & 21.6K & 43K & \xmark & \cmark & \cmark & \xmark & \xmark
& Detailed & 29K & 3.7K & 29K & 3.7K \\

SSG-VQA \cite{ssg_vqa} & \cmark & 1 & 25K & 960K & Endo & \cmark & \cmark & \xmark & \xmark
& Conv-1 & 29K & 3.7K & 273K & 35K \\

Surg-396K \cite{endochat} & \cmark & 3 & 41.4K & 396K & Endo & \xmark & \xmark & \cmark & \xmark
& Conv-2 & 29K & 3.7K & 251K & 32K \\

SurgVLM \cite{surgvlm} & \xmark & 16 & 1810K & 7798K & -- & \xmark & \xmark & -- & --
& Reas. & 29K & 3.7K & 170K & 22K \\

\textbf{MIS-VQA} & \cmark & 2 & 33K & 849K & \cmark & \xmark & \xmark & \cmark & \cmark
& \textbf{Total} & \textbf{29K} & \textbf{3.7K} & \textbf{751K} & \textbf{97K} \\
\hline
\multicolumn{15}{l}{$^{*}$Public availability refers to accessibility at the time of writing.} \\
\end{tabular}%
}
\end{table*}
\subsection{Comparison with Existing Datasets}

We compare MIS-VQA with existing surgical VQA benchmarks in Table~\ref{tab:dataset_comp_and_stats}. As illustrated in Fig.~\ref{fig:icl}(b), early datasets such as EndoVis-18-VQA and Cholec80-VQA rely heavily on template-based question generation, limiting linguistic diversity and enabling pattern memorization rather than true visual understanding. Moreover, existing Cholec-based VQA datasets, including Cholec80-VQA and Surg-396K, do not incorporate spatially grounded supervision, largely because ground-truth segmentation masks are sparsely available (e.g., only in CholecSeg8k). Consequently, spatial questions and spatially informed answers are generally absent, constraining the quality of captioning and scene-level reasoning. 

In contrast, MIS-VQA unifies concise and detailed captioning, hierarchical conversational grounding (Tier-1 and Tier-2), and explicit complex reasoning within a single framework. By augmenting sparse segmentation annotations with confidence-filtered predictions, MIS-VQA enables accurate spatially grounded supervision at scale while maintaining annotation reliability. This design provides broader supervision signals and supports more comprehensive and diagnostically targeted evaluation of multimodal surgical understanding. We make MIS-VQA publicly available to facilitate reproducible research and standardized benchmarking in surgical vision-language modeling.



\subsection{Results}

\begin{table*}[t]
\centering
\footnotesize
\setlength{\tabcolsep}{2.8pt}
\caption{Quantitative results on the EndoVis and Cholec parts of MIS-VQA across all response types. The best result in each column is shown in bold and the second best is underlined. COS: cosine similarity of sentence embeddings; MET: METEOR.}
\label{tab:results_endovis}
\resizebox{\textwidth}{!}{
\begin{tabular}{l|cccc|ccc|ccc|ccc|ccc}
\hline
 & \multicolumn{4}{c|}{Conv-1} & \multicolumn{3}{c|}{Conv-2} & \multicolumn{3}{c|}{Concise} & \multicolumn{3}{c|}{Detailed} & \multicolumn{3}{c}{Reasoning} \\
Model & GPT* & COS & BLEU-1 & BLEU-3 & GPT* & COS & MET & GPT* & COS & CIDEr & GPT* & COS & CIDEr & GPT* & COS & CIDEr \\
\hline
\multicolumn{17}{c}{\textbf{EndoVis}} \\
\hline
VisualBert \cite{visualbert} & 59.37 & 0.6023 & 27.87 & 16.39 & 41.98 & 0.7848 & 44.54 & -- & -- & -- & -- & -- & -- & -- & -- & -- \\
VisualBert RM \cite{endovis_vqa} & 50.70 & 0.5062 & 20.13 & 11.28 & 43.32 & 0.7780 & 43.49 & 21.85 & 0.8298 & 11.16 & -- & -- & -- & -- & -- & -- \\
EndoChat \cite{endochat} & 60.95 & \textbf{0.7295} & \textbf{54.06} & \textbf{42.57} & 51.60 & \underline{0.8690} & 56.84 & 24.32 & 0.8083 & 19.21 & 21.43 & 0.7048 & 1.05 & 70.76 & 0.8305 & 13.59 \\
LLaVA-SurgXpert (PeskaVLP) & 55.29 & 0.6758 & 47.87 & \underline{34.57} & 46.85 & 0.8565 & 55.79 & 22.71 & 0.8374 & 21.78 & 24.91 & 0.8604 & 4.27 & 76.60 & 0.8466 & 17.47 \\
LLaVA-SurgXpert (FASL) & 60.98 & \underline{0.6845} & \underline{48.27} & 34.37 & \underline{51.82} & 0.8673 & 57.20 & \textbf{36.17} & \underline{0.8897} & \textbf{40.29} & \textbf{39.98} & \underline{0.9015} & 6.71 & \textbf{78.47} & \textbf{0.8534} & \underline{19.74} \\
LLaVA-SurgXpert (SAM) & \textbf{62.51} & 0.6609 & 44.72 & 30.71 & \textbf{52.61} & \textbf{0.8711} & \textbf{57.85} & \underline{29.46} & \textbf{0.8940} & \underline{35.24} & \underline{31.47} & \textbf{0.9017} & \textbf{8.37} & \underline{78.44} & \underline{0.8527} & 19.51 \\
LLaVA-SurgXpert (CLIP) & \underline{61.06} & 0.6760 & 46.60 & 32.07 & 47.55 & 0.8626 & \underline{57.30} & 26.27 & 0.8843 & 34.10 & 28.53 & 0.8977 & \underline{8.34} & 77.16 & 0.8524 & \textbf{20.11} \\
\hline
\multicolumn{17}{c}{\textbf{Cholec}} \\
\hline
EndoChat \cite{endochat} & \textbf{73.15} & \textbf{0.8589} & \textbf{63.24} & \textbf{56.63} & \textbf{70.52} & \textbf{0.8641} & \textbf{63.76} & 47.55 & 0.7925 & 32.33 & 46.75 & 0.8583 & 8.83 & 80.47 & 0.8688 & 20.79 \\
LLaVA-SurgXpert (SAM) & \underline{66.99} & \underline{0.8528} & \underline{62.86} & \underline{55.84} & \underline{65.37} & \underline{0.8579} & \underline{63.11} & \textbf{52.05} & \textbf{0.8197} & \textbf{60.73} & \textbf{51.60} & \textbf{0.8771} & \textbf{13.03} & \textbf{81.92} & \textbf{0.8761} & \textbf{23.70} \\
LLaVA-SurgXpert (FASL) & 59.57 & 0.8429 & 61.40 & 54.16 & 55.54 & 0.8274 & 59.16 & 41.04 & 0.7528 & 35.91 & 39.11 & 0.8291 & 5.02 & 80.83 & 0.8743 & 23.29 \\
LLaVA-SurgXpert (CLIP) & 64.44 & 0.8499 & 62.60 & 55.50 & 62.75 & 0.8516 & 62.44 & \underline{48.45} & \underline{0.8030} & \underline{55.38} & \underline{47.56} & \underline{0.8688} & \underline{11.55} & \underline{81.42} & \underline{0.8760} & \underline{23.45} \\
\hline
\multicolumn{17}{l}{$^{*}$GPT score: both predicted and GT answers are scored by an LLM judge conditioned on the annotation-driven context; the normalized value is reported.} \\
\end{tabular}
}
\end{table*}
\textbf{Overall performance trends.}
Table~\ref{tab:results_endovis} summarizes the quantitative results on the EndoVis and Cholec subsets of MIS-VQA. Across both procedures and all response types, the LLaVA-SurgXpert family consistently outperforms earlier surgical VQA baselines such as VisualBert and VisualBert-RM.  EndoChat remains the strongest baseline, particularly for conversational tasks on both procedures, reflecting its design around dialogue-oriented supervision. However, its advantage diminishes on more complex outputs such as detailed description and reasoning, where LLaVA-SurgXpert variants show clear gains. Importantly, no single encoder dominates across all categories; instead, encoder rankings vary depending on the response type. This variability highlights the benefit of the hierarchical evaluation protocol in MIS-VQA, which exposes different strengths in perceptual grounding, caption generation, and compositional reasoning.

\textbf{Impact of segmentation-aware encoders.} 
Encoders pretrained with spatial supervision demonstrate clear advantages. On EndoVis, segmentation-aware encoders SAM and FASL consistently rank first and second across most response types. SAM achieves the highest GPT-Scores on conversational tasks, while FASL leads the concise, detailed captioning and complex reasoning tasks. These results suggest that segmentation pretraining provides strong geometric inductive biases that improve visual grounding in surgical scenes. On Cholec, LLaVA-SurgXpert (SAM) achieves the strongest overall performance, reaching the highest reasoning GPT-Score of 81.92 and CIDEr score of 23.70.

\textbf{Role of GPT-based scoring.}
While standard metrics such as BLEU and METEOR measure lexical overlap with the ground truth, the normalized GPT-Score provides a more holistic estimate of answer quality. This is evident in the EndoVis Conv-1 results where EndoChat achieves the highest BLEU-1 and BLEU-3 scores (54.06) but a lower GPT-Score (60.95) than LLaVA-SurgXpert (SAM), which achieves the best GPT-Score of 62.51. This discrepancy suggests that high n-gram overlap does not necessarily correspond to stronger visual grounding or clinical correctness. Given the design of the utilized GPT evaluation which conditions on the annotation-derived context, it provides a more reliable indicator of multimodal reasoning quality than single-reference metrics.

\textbf{Influence of pretraining objectives.}
Interestingly, domain specialization alone does not guarantee superior performance. LLaVA-SurgXpert with PeskaVLP consistently ranks lowest among the tested encoders on EndoVis. This behavior may be attributed to its ResNet-based architecture, which produces coarser spatial feature maps compared with ViT-based encoders, and to its pretraining objective that emphasizes video-level procedural semantics rather than dense spatial correspondence.

\section{Conclusion}
In this work, we addressed limitations of existing surgical vision-language datasets, including limited scale, template-driven supervision, and restricted spatial grounding due to sparse or incomplete segmentation annotations, which hinder structured evaluation across perceptual grounding and compositional reasoning. We introduced MIS-VQA, a large-scale dataset constructed through a structured annotation-to-language pipeline that leverages segmentation-informed context to enable consistent spatially grounded captions and compositional reasoning. Together with LLaVA-SurgXpert, a family of surgical vision-language models used as a controlled framework to analyze the impact of visual encoder pretraining on multimodal surgical understanding, this work establishes a unified benchmark and analysis framework for surgical vision-language modeling.  Future directions include clinical validation, extension to video-based VQA for temporal reasoning, improved segmentation reliability through finer-grained or ensemble-based filtering, and scaling to larger multimodal models to enhance robustness and reasoning capacity.

\subsubsection{Acknowledgements} Research reported in this publication was supported by the Qatar Research Development and Innovation Council (QRDI) grant number ARG01-0522-230266. Disclaimer:  The content is solely the responsibility of the authors and does not necessarily represent the official views of Qatar Research Development and Innovation Council.

%
%
%
\bibliographystyle{splncs04}
\bibliography{references}

@inproceedings{hamdy2026decoupling,
  title     = {Decoupling Vision and Reasoning: A Data-Efficient Pipeline for Surgical {VQA}},
  author    = {Hamdy, Mohamed and Ahmed, Fatimaelzahraa Ali and Abdel-Ghani, Muraam and Arsalan, Muhammad and Suganthan, Ponnuthurai Nagaratnam and Al-Jalham, Khalid and Al-Ali, Abdulaziz and Balakrishnan, Shidin},
  booktitle = {Medical Imaging with Deep Learning},
  year      = {2026}
}

@inproceedings{midlshort,
  title     = {Segmentation-Informed Captioning: A Multi-Stage Pipeline for Surgical Vision--Language Dataset Generation},
  author    = {Hamdy, Mohamed and Ahmed, Fatimaelzahraa Ali and Ahmed, Mariam and AbuHaweeleh, Mohannad Natheef and Abdel-Ghani, Muraam and Arsalan, Muhammad and Al-Ali, Abdulaziz and Balakrishnan, Shidin},
  booktitle = {Medical Imaging with Deep Learning -- Short Papers},
  year      = {2025}
}

@inproceedings{fasl,
  author    = {Abdel-Ghani, Muraam and Mahmoud, Ali and Mohamed, Ali and Ahmed, Fatmaelzahraa and Mohamed, Arsalan and Al-Ali, Abdulaziz and Balakrishnan, Shidin},
  title     = {{FASL-Seg}: Anatomy and Tool Segmentation of Surgical Scenes},
  booktitle = {ECAI 2025},
  year      = {2025}
}

@inproceedings{endovis_vqa,
  title     = {{Surgical-VQA}: Visual Question Answering in Surgical Scenes Using Transformer},
  author    = {Seenivasan, Lalithkumar and Islam, Mobarakol and Krishna, Adithya K. and Ren, Hongliang},
  booktitle = {Medical Image Computing and Computer Assisted Intervention -- MICCAI 2022},
  pages     = {33--43},
  year      = {2022}
}

@article{ssg_vqa,
  title   = {Advancing Surgical {VQA} with Scene Graph Knowledge},
  author  = {Yuan, Kun and Kattel, Manasi and Lavanchy, Joel L. and Navab, Nassir and Srivastav, Vinkle and Padoy, Nicolas},
  journal = {International Journal of Computer Assisted Radiology and Surgery},
  volume  = {19},
  number  = {7},
  pages   = {1409--1417},
  year    = {2024}
}

@article{surgvlm,
  title   = {{SurgVLM}: A Large Vision-Language Model and Systematic Evaluation Benchmark for Surgical Intelligence},
  author  = {Zeng, Zhitao and Zhuo, Zhu and Jia, Xiaojun and Zhang, Erli and Wu, Junde and Zhang, Jiaan and Wang, Yuxuan and Low, Chang Han and Jiang, Jian and Zheng, Zilong and others},
  journal = {arXiv preprint arXiv:2506.02555},
  year    = {2025}
}

@article{peskavlp,
  title   = {Procedure-Aware Surgical Video-Language Pretraining with Hierarchical Knowledge Augmentation},
  author  = {Yuan, Kun and Navab, Nassir and Padoy, Nicolas and others},
  journal = {Advances in Neural Information Processing Systems},
  volume  = {37},
  pages   = {122952--122983},
  year    = {2024}
}

@article{endovis18,
  title   = {2018 Robotic Scene Segmentation Challenge},
  author  = {Allan, Max and Kondo, Satoshi and Bodenstedt, Sebastian and Leger, Stefan and Kadkhodamohammadi, Rahim and Luengo, Imanol and Fuentes, Felix and Flouty, Evangello and Mohammed, Ahmed and Pedersen, Marius and others},
  journal = {arXiv preprint arXiv:2001.11190},
  year    = {2020}
}

@inproceedings{inst_tissue1,
  title     = {Learning and Reasoning with the Graph Structure Representation in Robotic Surgery},
  author    = {Islam, Mobarakol and Seenivasan, Lalithkumar and Ming, Lim Chwee and Ren, Hongliang},
  booktitle = {Medical Image Computing and Computer Assisted Intervention -- MICCAI 2020},
  pages     = {627--636},
  year      = {2020}
}

@article{inst_tissue2,
  title   = {Global-Reasoned Multi-Task Learning Model for Surgical Scene Understanding},
  author  = {Seenivasan, Lalithkumar and Mitheran, Sai and Islam, Mobarakol and Ren, Hongliang},
  journal = {IEEE Robotics and Automation Letters},
  volume  = {7},
  number  = {2},
  pages   = {3858--3865},
  year    = {2022}
}

@article{cv_surgery,
  title   = {Computer Vision in Surgery: From Potential to Clinical Value},
  author  = {Mascagni, Pietro and Alapatt, Deepak and Sestini, Luca and Altieri, Maria S. and Madani, Amin and Watanabe, Yusuke and Alseidi, Adnan and Redan, Jay A. and Alfieri, Sergio and Costamagna, Guido and others},
  journal = {npj Digital Medicine},
  volume  = {5},
  number  = {1},
  pages   = {163},
  year    = {2022}
}

@article{lancet,
  title   = {Laparoscopic Versus Open Parenchymal Preserving Liver Resections in the Posterosuperior Segments ({ORANGE} Segments): A Multicentre, Single-Blind, Randomised Controlled Trial},
  author  = {Sijberden, Jasper P. and Kuemmerli, Christoph and Ratti, Francesca and D'Hondt, Mathieu and Sutcliffe, Robert P. and Troisi, Roberto I. and Efanov, Mikhail and Fichtinger, Robert S. and D{\'\i}az-Nieto, Rafael and Ettorre, Giuseppe M. and others},
  journal = {The Lancet Regional Health--Europe},
  volume  = {51},
  year    = {2025}
}

@article{laparoscopic_review,
  title   = {Laparoscopic Surgery: A Qualified Systematic Review},
  author  = {Buia, Alexander and Stockhausen, Florian and Hanisch, Ernst},
  journal = {World Journal of Methodology},
  volume  = {5},
  number  = {4},
  pages   = {238},
  year    = {2015}
}

@article{cholecseg8k,
  title   = {{CholecSeg8k}: A Semantic Segmentation Dataset for Laparoscopic Cholecystectomy Based on {Cholec80}},
  author  = {Hong, W.-Y. and Kao, C.-L. and Kuo, Y.-H. and Wang, J.-R. and Chang, W.-L. and Shih, C.-S.},
  journal = {arXiv preprint arXiv:2012.12453},
  year    = {2020}
}

@article{rendezvous_triplet,
  title   = {Rendezvous: Attention Mechanisms for the Recognition of Surgical Action Triplets in Endoscopic Videos},
  author  = {Nwoye, Chinedu Innocent and Yu, Tong and Gonzalez, Cristians and Seeliger, Barbara and Mascagni, Pietro and Mutter, Didier and Marescaux, Jacques and Padoy, Nicolas},
  journal = {Medical Image Analysis},
  volume  = {78},
  pages   = {102433},
  year    = {2022}
}

@article{endonet,
  title   = {{EndoNet}: A Deep Architecture for Recognition Tasks on Laparoscopic Videos},
  author  = {Twinanda, Andru P. and Shehata, Sherif and Mutter, Didier and Marescaux, Jacques and De Mathelin, Michel and Padoy, Nicolas},
  journal = {IEEE Transactions on Medical Imaging},
  volume  = {36},
  number  = {1},
  pages   = {86--97},
  year    = {2017}
}

@article{visualbert,
  title   = {{VisualBERT}: A Simple and Performant Baseline for Vision and Language},
  author  = {Li, Liunian Harold and Yatskar, Mark and Yin, Da and Hsieh, Cho-Jui and Chang, Kai-Wei},
  journal = {arXiv preprint arXiv:1908.03557},
  year    = {2019}
}

@article{Deep_Learning_Sys,
  author  = {Ahmed, F. A. and Yousef, M. and Ahmed, M. A. and Ali, H. O. and Mahboob, A. and Ali, H. and Shah, Z. and Aboumarzouk, O. and Al Ansari, A. and Balakrishnan, S.},
  title   = {Deep Learning for Surgical Instrument Recognition and Segmentation in Robotic-Assisted Surgeries: A Systematic Review},
  journal = {Artificial Intelligence Review},
  year    = {2024}
}

@article{endochat,
  title   = {{EndoChat}: Grounded Multimodal Large Language Model for Endoscopic Surgery},
  author  = {Wang, Guankun and Bai, Long and Wang, Junyi and Yuan, Kun and Li, Zhen and Jiang, Tianxu and He, Xiting and Wu, Jinlin and Chen, Zhen and Lei, Zhen and Liu, Hongbin and Wang, Jiazheng and Zhang, Fan and Padoy, Nicolas and Navab, Nassir and Ren, Hongliang},
  journal = {Medical Image Analysis},
  volume  = {107},
  pages   = {103789},
  year    = {2026}
}

@article{llava,
  title   = {Visual Instruction Tuning},
  author  = {Liu, Haotian and Li, Chunyuan and Wu, Qingyang and Lee, Yong Jae},
  journal = {Advances in Neural Information Processing Systems},
  volume  = {36},
  pages   = {34892--34916},
  year    = {2023}
}

%




\end{document}